%% file: 00_main.tex
\documentclass[letterpaper, 10 pt, conference]{ieeeconf}
\IEEEoverridecommandlockouts
\usepackage{mathtools}
\usepackage{cite}
\usepackage[dvipsnames, table]{xcolor}
\usepackage{amsmath,amssymb,amsfonts}
\usepackage[font=small, labelfont=bf]{caption}
\usepackage[ruled, lined, linesnumbered, commentsnumbered, longend, noend]{algorithm2e}
\usepackage{hyperref}
\usepackage{algorithmic}
\usepackage{graphicx}
\usepackage{textcomp}
\usepackage{xcolor}
\usepackage{breqn}
\usepackage{gensymb}
\usepackage{cuted}
\usepackage{capt-of}
\usepackage{booktabs}
\usepackage{array}
\usepackage{diagbox}
\usepackage{pifont}
\usepackage{multirow}
\usepackage{makecell}
\usepackage{bm}

\newcommand{\cmark}{\ding{51}}
\newcommand{\xmark}{\textcolor{gray}{\ding{55}}}
\newcommand{\omark}{\textcolor{gray}{\ding{108}}}

\newtheorem{theorem}{Theorem}[section]

\newtheorem{lemma}[theorem]{Lemma}

\usepackage{fancyhdr} 
\def\BibTeX{{\rm B\kern-.05em{\sc i\kern-.025em b}\kern-.08em
    T\kern-.1667em\lower.7ex\hbox{E}\kern-.125emX}}

\definecolor{pink}{RGB}{255, 192, 203}

\definecolor{darkgreen}{RGB}{83, 199, 34}

\title{\LARGE \bf \textit{TD-GRPC}: Temporal Difference Learning with Group Relative Policy Constraint for Humanoid Locomotion}

\author{Khang Nguyen$^{1}$, Khai Nguyen$^{2}$, An T. Le$^{3}$, Jan Peters$^{3,4,5}$, Manfred Huber$^{1}$, Vien Ngo$^{2}$, and Minh Nhat Vu$^{6}$
\thanks{$^{1}$University of Texas at Arlington, Texas, USA}
\thanks{$^{2}$VinRobotics, Hanoi, Vietnam}
\thanks{$^{3}$Intelligent Autonomous Systems Lab, TU Darmstadt, Germany} 
\thanks{$^{4}$German Research Center for AI (DFKI), SAIROL, Darmstadt, Germany} 
\thanks{$^{5}$Hessian.AI, Darmstadt, Germany}
\thanks{$^{6}$Automation \& Control Institute (ACIN), TU Wien, Vienna, Austria} 
\thanks{E-mails: \href{mailto:khang.nguyen8@mavs.uta.edu}{\text{khang.nguyen8@mavs.uta.edu}}, \href{mailto:minh.vu@ait.ac.at}{\text{minh.vu@ait.ac.at}}.}
}

\begin{document}

\maketitle
\thispagestyle{empty}
\pagestyle{empty}

\begin{strip}
    \vspace{-85pt}
    \centering
    \includegraphics[width=1.00\linewidth]{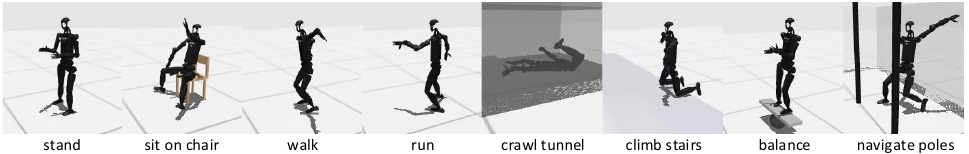}
    \vspace{-14pt}
    \captionof{figure}{\textbf{Locomotion Tasks Performed by the Unitree H1-2 Humanoid with TD-GRPC:} standing, sitting on a chair, walking, running, crawling under a tunnel, climbing stairs, balancing on a ball-board toy, and navigating through standing poles while avoiding collision.}
    \vspace{-12pt}
    \label{fig:teaser}
\end{strip}

\begin{abstract}
    Robot learning in high-dimensional control settings, such as humanoid locomotion, presents persistent challenges for reinforcement learning (RL) algorithms due to unstable dynamics, complex contact interactions, and sensitivity to distributional shifts during training. Model-based methods, \textit{e.g.}, Temporal-Difference Model Predictive Control (TD-MPC), have demonstrated promising results by combining short-horizon planning with value-based learning, enabling efficient solutions for basic locomotion tasks. However, these approaches remain ineffective in addressing policy mismatch and instability introduced by off-policy updates. Thus, in this work, we introduce Temporal-Difference Group Relative Policy Constraint (TD-GRPC), an extension of the TD-MPC framework that unifies Group Relative Policy Optimization (GRPO) with explicit Policy Constraints (PC). TD-GRPC applies a trust-region constraint in the latent policy space to maintain consistency between the planning priors and learned rollouts, while leveraging group-relative ranking to assess and preserve the physical feasibility of candidate trajectories. Unlike prior methods, TD-GRPC achieves robust motions without modifying the underlying planner, enabling flexible planning and policy learning. We validate our method across a locomotion task suite ranging from basic walking to highly dynamic movements on the 26-DoF Unitree H1-2 humanoid robot. Through simulation results, TD-GRPC demonstrates its improvements in stability and policy robustness with sampling efficiency while training for complex humanoid control tasks.
\end{abstract}

\input{01_introduction}
\input{02_related_work}
\input{03_methodology}
\input{04_evaluation}
\input{05_conclusions}

\bibliographystyle{IEEEtran}
\bibliography{references}

\end{document}

%% file: 01_introduction.tex
\section{Introduction}

Humanoid locomotion is one of the most challenging domains in control and reinforcement learning (RL), due to its inherently high-dimensional and dynamically complex body structure \cite{radosavovic2023learning, radosavovic2024real}. Achieving robust and adaptive control to accomplish locomotion tasks requires an algorithm that can not only plan effectively in the face of environmental uncertainty but also learn generalizable behaviors from limited iterations. While model predictive control (MPC) offers strong short-term decision-making capabilities via real-time trajectory optimization, its effectiveness is often bounded by model accuracy alongside the need for meticulously hand-crafted cost functions. In contrast, RL enables more flexible, data-driven behavior learning but typically suffers from sampling inefficiency and unstable policy updates, especially for model-based RL (MBRL) approaches \cite{nagabandi2018neural, chua2018deep, argenson2020model}. 

\begin{table*}[t]
    \vspace{5pt}
    \centering
    \caption{Comparison of policy learning- and optimization-based attributes across prior model-free and model-based approaches, including MPC \cite{kouvaritakis2016model}, SAC \cite{haarnoja2018soft}, MuZero \cite{schrittwieser2020mastering}, EfficientZero \cite{ye2021mastering}, LOOP \cite{sikchi2022learning}, TD-MPC2 \cite{hansen2023td}, TD-M(PC)$^2$ \cite{lin2025td}, and our proposed method. The highlighted columns introduce policy constraints and group-relative rankings, which are critical for further policy optimization of humanoid locomotion.}
    \begin{tabular}{l | ccccccc}
        \toprule
        \diagbox{Method}{Attr.} & \makecell{Continuous \\ controller} & \makecell{Model learning \\ objective} & \makecell{Value function \\ learning} & \cellcolor{gray!20}\makecell{Policy \\ constraint} & \cellcolor{gray!20}\makecell{Group-relative \\ advantage} & \makecell{Inference \\ strategy} & \makecell{Computation \\ cost} \\
        \midrule \midrule
        MPC \cite{kouvaritakis2016model} & \cmark & not defined & \xmark & \cellcolor{gray!20}\xmark & \cellcolor{gray!20}\xmark & CEM & High \\
        SAC \cite{haarnoja2018soft} & \cmark & not defined & \cmark & \cellcolor{gray!20}\xmark & \cellcolor{gray!20}\xmark & policy & Low \\
        \midrule
        MuZero \cite{schrittwieser2020mastering} & \xmark & reward + value & \cmark & \cellcolor{gray!20}\xmark & \cellcolor{gray!20}\xmark & MCTS-based policy & Moderate \\
        EfficientZero \cite{ye2021mastering} & \xmark & reward + value & \cmark & \cellcolor{gray!20}\xmark & \cellcolor{gray!20}\xmark & MCTS-based policy & Moderate \\
        LOOP \cite{sikchi2022learning} & \cmark & state & \cmark & \cellcolor{gray!20}\xmark & \cellcolor{gray!20}\xmark & CEM-based policy & Moderate \\
        \midrule
        TD-MPC2 \cite{hansen2023td} & \cmark & reward + value + state & \cmark & \cellcolor{gray!20}\xmark & \cellcolor{gray!20}\xmark & CEM-based policy & Low \\
        TD-M(PC)$^2$ \cite{lin2025td} & \cmark & reward + value + state & \cmark & \cellcolor{gray!20}\cmark & \cellcolor{gray!20}\xmark & CEM-based policy & Low \\
        \midrule
        \textbf{TD-GRPC (Ours)} & \cmark & reward + value + state & \cmark & \cellcolor{gray!20}\cmark & \cellcolor{gray!20}\cmark & CEM-based policy & Low \\
        \bottomrule
    \end{tabular}
    \label{tab:attributes}
    \vspace{-12pt}
\end{table*}

While MBRL holds great promises for sample efficiency by leveraging learned dynamics models for planning and value estimation \cite{sutton1991dyna, janner2019trust}, it often suffers from compounding model errors and planning biases in games \cite{kaiser2019model, schrittwieser2020mastering, ye2021mastering, pham2025pay}, in quadruped robots control \cite{hwangbo2019learning, lee2020learning, miki2022learning}, and particularly in high-dimensional tasks such as in humanoid control \cite{lin2025td}. Offline approaches to MBRL have attempted to mitigate these limitations by decoupling data collection from learning \cite{argenson2020model, janner2019trust}. Additionally, PlaNet \cite{hafner2019dream} and Dreamer variants \cite{hafner2019dream, hafner2020mastering, hafner2023mastering} have explored latent dynamics for improving policy learning from pixels \cite{hafner2019dream, ha2018recurrent}. Nevertheless, many of these methods still exhibit weak generalization and lack the temporal abstraction necessary for robust and trustworthy long-horizon reasoning \cite{chua2018deep, lambert2020objective, xu2022feasibility}.

Earlier works have attempted to merge planning and learning more tightly. Temporal-Difference Model Predictive Control (TD-MPC) \cite{hansen2022temporal} addresses these issues by training latent dynamics models jointly with value functions using TD learning, avoiding the pitfalls of purely supervised model learning and enabling better credit assignment over long horizons. TD-MPC2 \cite{hansen2023td} further refines this framework by introducing stabilized actor-critic updates in the latent space, achieving improved consistency in off-policy learning. However, challenges persist despite these improvements due to the mismatch between training targets and planner-induced policies, destabilizing off-policy MBRL \cite{lambert2020objective, clavera2020model}. Moreover, the lack of constraints during policy improvement steps can result in aggressive updates that cause significant distribution shifts, as studied in \cite{fujimoto2019off, kostrikov2021offline, bellemare2017distributional, levine2020offline}.

To bridge this gap, recent advancements have demonstrated promising results by combining the planning efficiency of MPC with the adaptability of TD learning. Their variants extend this hybrid paradigm by incorporating latent dynamics models and stabilized value estimation, enabling more scalable and robust control in continuous settings. However, key challenges remain unresolved, particularly the mismatch between learned policy rollouts and the TD targets used in off-policy training, leading to degraded performance and instability over time and failing to accomplish the tasks.

In this work, we propose to alleviate these limitations by extending the TD-MPC framework \cite{hansen2023td} with Group Relative Policy Optimization (GRPO) \cite{shao2024deepseekmath} to improve stability and sample efficiency in humanoid locomotion settings. Inspired by recent state-of-the-art techniques in constrained policy optimization, our method further combines a trust-region constraint directly on the policy prior in the latent space, preserving planning feasibility while mitigating distributional drift during learning. In addition, our technique achieves a faster convergence rate than other baselines.  
Our contributions are summarized as follows:
\begin{itemize}
    \item We propose a TD-GRPC framework, which integrates GRPO with explicit trust-region constraints in the latent space. In addition, we provide theoretical insights into its supporting role in stabilizing off-policy MBRL.
    \item We empirically validate TD-GRPC on the locomotion suite of \texttt{HumanoidBench} \cite{sferrazza2024humanoidbench}, demonstrating significant improvements in sample efficiency and policy robustness across a set of humanoid locomotion tasks.
\end{itemize}

%% file: 02_related_work.tex
\section{Related Work}

\begin{figure*}[t]
    \vspace{4pt}
    \centering
    \includegraphics[width=1.0\linewidth]{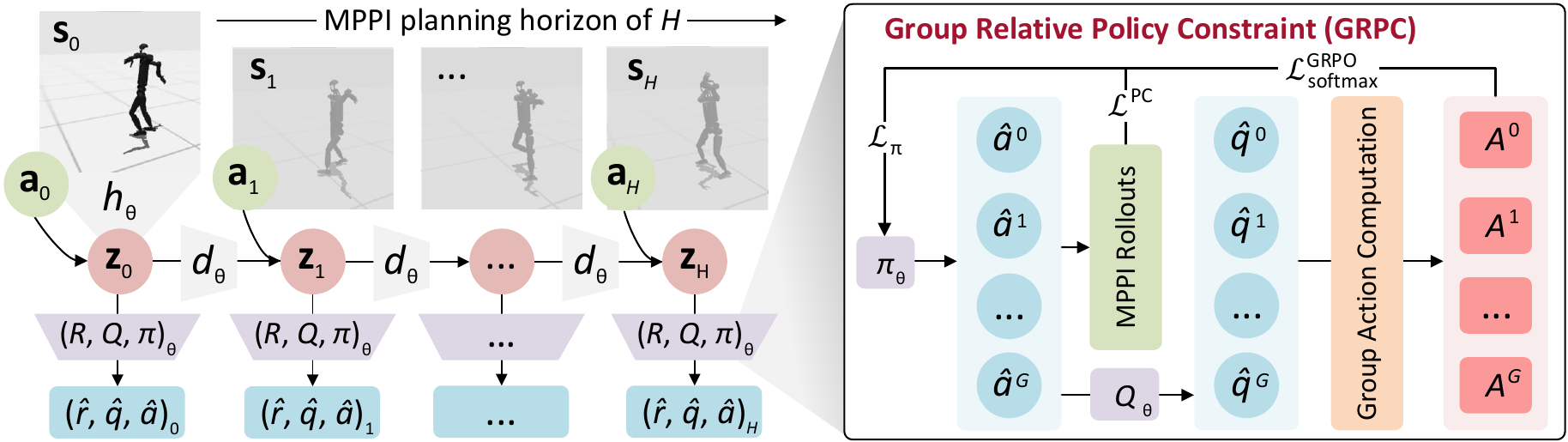}
    \caption{\textbf{Overview of TD-GRPC for Humanoid Locomotion:} Starting from an initial state $\textbf{s}_0$ encoded into latent state $\textbf{z}_0$ with an encoder $h_{\theta}$, a latent dynamics model $d_{\theta}$ takes an action $\textbf{a}_t$ and the latent state $\textbf{z}_t$ to predict the next latent state $\textbf{z}_{t+1}$ across $H$ steps of MPPI planning horizon. In each step, the reward, $Q$-value, and action are estimated via the MLPs $R_{\theta}$, $Q_{\theta}$, and $\pi_{\theta}$, enabling latent-space planning with TD targets. At each state, sampled groups of actions are rolled out to evaluate $Q$-values, which are used to compute softmax-based advantage scores $A^{g}$ of the $g^{th}$ group. These scores are then used in the GRPC objective to guide the policy toward high-value actions while minimizing variance. To prevent excessive policy shifts, a trust-region constraint is imposed via a KL divergence penalty between the current and a prior MPPI-derived policies, enforcing residual learning with bounded policy divergence.}
    \vspace{-16pt}
    \label{fig:tdgrpc_framework}
\end{figure*}

\textbf{Temporal-Difference Model Predictive Control:} Humanoid locomotion poses unique challenges in control due to its high-dimensional continuous action spaces, unstable dynamics, and complex contact behaviors from the environment \cite{peters2003reinforcement}. TD-MPC has demonstrated its potential to address these difficulties by combining the short-horizon planning strategies of MPC-based approaches alongside RL's sample-efficient, value-driven adaptability. Through this line of research, notable works \cite{sikchi2022learning, hansen2022temporal, hansen2023td} have revealed that incorporating TD learning into MPC-based methods allows for learning flexible value functions without the need for meticulously designed cost functions. In particular, TD-MPC2 \cite{hansen2023td} extends the vanilla TD-MPC framework \cite{hansen2022temporal} by learning more scalable latent world models for continuous control, mitigating compounding model errors, and stabilizing planning. These insights are critical for humanoid locomotion, where model inaccuracies can rapidly destabilize gaits. Thus, by integrating TD learning with MPC-styled control paradigms, modern frameworks enhance sample efficiency, robustness to modeling errors, and adaptability to high-dimensional motor tasks, making them well-suited for complex humanoid behaviors. Our work further extends by algorithmically improving stability and data efficiency through GRPO, similar to language models, and with an additional explicit trust-region constraint for rolled-out policies from the latent representations.

\textbf{Policy Constraint for Off-Policy Learning:} Policy mismatching for robot learning, especially for humanoid locomotion, remains a fundamental challenge in off-policy learning, which is highly exposed to shifts between the learned rolled-out actions and TD targets. Moreover, the accumulation and compound of bootstrapping errors further induce cumulative errors and poor generalization \cite{kumar2019stabilizing, kumar2020conservative}. Offline-policy RL methods have greatly improved policy learning from fixed datasets. Previous works in this scope handled the distributional shifts by explicitly regularizing the learned policies towards the expert policies \cite{kumar2019stabilizing, fujimoto2021minimalist}. In contrast, others \cite{fujimoto2019off, peng2019advantage} employ importance sampling techniques to correct for out-of-distribution queries. Alternative approaches, such as in-sample learning \cite{garg2023extreme, kostrikov2021offline}, recover reliable policies implicitly by restricting updates to observed actions, bypassing explicit distribution constraints. The off-policy challenge is also critical in MBRL when leveraging policy or value priors for planning. For instance, LOOP \cite{sikchi2022learning} proposes actor regularization control, introducing a conservatism mechanism during planning to stabilize online learning. Unlike these methods, our approach enforces distributional constraints directly on the policy prior without altering the planner, enabling greater flexibility in planning while preserving stability and fast policy updates.

To recap, we compare attributes between our proposed method and prior model-free and model-based approaches in Tab. \ref{tab:attributes} on policy constraint and optimization alongside model learning objectives, value function learning, and use cases.

%% file: 03_methodology.tex
\section{Temporal Difference Learning with Group Relative Policy Constraint}
\label{sec:methodology}

\subsection{Problem Formulation}
Any humanoid locomotion task can be modeled as an infinite-horizon Markov Decision Process (MDP) characterized as $\mathcal{M} = ( \mathcal{S}, \mathcal{A}, \mathcal{P}, r, \gamma )$, with $\mathcal{S}$ is the state space, $\mathcal{A}$ is the action space, $\mathcal{P}: \mathcal{S} \times \mathcal{A} \rightarrow \mathcal{S}$ is the dynamics function, $r: \mathcal{S} \times \mathcal{A} \rightarrow \mathbb{R}$ is the reward function, and $\gamma \in (0, 1]$ is the discount factor. The problem is to learn the parameters $\theta$ for the policy network $\Pi_{\theta}: \mathcal{S} \rightarrow \mathcal{A}$ that continuously controls the robot with optimal state-based actions by maximizing the discounted cumulative rewards along a trajectory $\Gamma$:
\begin{equation}
    J^{\pi} = \mathbb{E}_{\Gamma \sim \Pi_{\theta}} \left[ \sum_{t=0}^{\infty} \gamma^{t} r(\textbf{s}_{t}, \textbf{a}_{t}) \right],
    \label{eq:objective}
\end{equation}
where $\Gamma = (\textbf{s}_{t}, \textbf{a}_{t})$ with each action $\textbf{a}_{t}$ is sampled from the policy network $\Pi_{\theta}(\textbf{s}_{t})$, and $\textbf{s}_{t} = \mathcal{P}(\textbf{s}_{t-1}, \textbf{a}_{t-1})$ is the transitional state based on the previous state and action.

Traditional model-free methods primarily focus on learning this policy directly, but they usually require abundant training data. On the other hand, model-based methods can be more sample-efficient as a learned dynamics model is used to simulate outcomes. Still, they often struggle with planning over long horizons due to inaccuracies in the model and high computation costs. Due to these, Hasen \textit{et al.} \cite{hansen2022temporal, hansen2023td} proposed the integration of a sampling-based MPC method, such as MPPI \cite{williams2016aggressive}, as a local trajectory optimizer for short-horizon planning using a learned latent dynamics model, and then extended it with a value function that predicts future returns. Action sequences of length $H$ are sampled as latent trajectories generated by the learned dynamics model, and estimate the total return $\phi_{\Gamma}$ of a sampled trajectory $\Gamma$:
\begin{equation}
    \phi_{\Gamma} = \mathbb{E}_{\Gamma} \left[ \gamma^{H} Q_{\theta} (\textbf{z}_{H}, \textbf{a}_{H}) + \sum_{t=0}^{H-1} \gamma^{t} R_{\theta}(\textbf{z}_{t}, \textbf{a}_{t}) \right],
    \label{eq:traj_update}
\end{equation}
where $\textbf{z}_{t} = h_{\theta}(\textbf{s}_{t})$ is the latent representation that selectively captures the relevant dynamics of the state $\textbf{s}_{t}$, rather than all observation dimensions, $\textbf{z}_{t} = d_{\theta}(\textbf{s}_{t-1}, \textbf{a}_{t-1})$ represents the next latent representation under the latent dynamics model $d_{\theta}$, $\hat{r}_{t} = R_{\theta} (\textbf{s}_{t}, \textbf{a}_{t})$ and $\hat{q}_{t} = Q_{\theta}(\textbf{s}_{t}, \textbf{a}_{t})$ denote the predicted reward and value under MLPs with $ \textbf{a}_{t} \sim \mathcal{N}(\mu_{t}, \sigma_{t})$ describes the trajectory distributions in Eq. \ref{eq:traj_update}, which are expressed as:
\begin{equation}
    \mu_{t} = \frac{\sum_{i=1}^k \Omega_i \Gamma_i^*}{\sum_{i=1}^k \Omega_i}, \quad \sigma_{t}^{2} = \frac{\sum_{i=1}^k \Omega_i(\Gamma_i^* - \mu^j)^2}{\sum_{i=1}^k \Omega_i} ,
    \label{eq:traj_moments}
\end{equation}
where $\Omega_i = \exp(\tau \phi_{\Gamma,i}^*)$, $\tau$ is a temperature parameter, and $\Gamma_i^*$ denotes the $i^{\text{th}}$ of top-$k$ trajectory corresponding to return estimate $\phi_{\Gamma}^*$.  In the RL-based robot learning context, Eq. \ref{eq:traj_update} can therefore be called as an $H$-step look-ahead policy, where the RL-based planner iteratively maximizes the cost (reward) of the first step of the planning horizon until the learning objective for locomotion is accomplished.

\subsection{TD Learning with Group Relative Policy Constraint}

\subsubsection{Policy Constraint as Residual Learning}
In TD-MPC2 \cite{hansen2023td}, the value function is learned using approximate policy iteration, similar to conventional off-policy RL methods. To further understand how various sources of approximation error influence the overall performance of the $H$-step look-ahead planner policy, we look into Lemma \ref{lemma:singh_yee} from \cite{singh1994upper}, which characterizes the asymptotic sub-optimality of the planner-induced policy $\mu_{k}$ in terms of the value approximation error of the policy $\pi_{k}$, the model prediction error in total variation distance, and the planner sub-optimality.
\begin{lemma}[\textit{Singh and Yee} \cite{singh1994upper}]
    \label{lemma:singh_yee}
    Suppose we have a value function $V^{\pi}(\textbf{s}) = \mathbb{E}_\pi \left[\sum_{t=0}^{\infty} \gamma^t r(\textbf{s}_t, \textbf{a}_t) \mid \textbf{s}_0 = \textbf{s} \right]$ such that $\max_s|V^*(\textbf{s}) - \hat{V}(\textbf{s})| \leq \varepsilon_v$. The performance of the $1$-step greedy policy $\pi_{\hat{V}}$ is bounded by:
    \begin{equation*}
        \frac{1 - \gamma}{2 \gamma} \left| J^{\pi^*} - J^{\pi_{\hat{V}}} \right| \leq \varepsilon_v
    \end{equation*}
\end{lemma}

\setlength{\textfloatsep}{4pt}
\begin{algorithm}[t]
    \caption{TD-GRPC}
    \label{alg:td_grpc}
    \begin{normalsize}
        \DontPrintSemicolon
        \SetKwInOut{KwIn}{Input}
        \SetKwInOut{KwOut}{Output}
        \SetKwFunction{FMainTrain}{td\_grpc\_training}
        \SetKwProg{Pn}{function}{}{}
        \KwIn{$T:$ trajectory length, $H$: planning horizon of MPPI, $G$: number of groups, $S$: number of iterations, $\mathcal{D}$: latent buffer} 
        \Pn{\FMainTrain{$T$, $H$, $G$, $S$, $\mathcal{D}$}}{
            \While{training}{
                \textcolor{gray}{// collecting trajs using latent-space MPPI} \\
                \For{$t = 0, \dots, T$}{
                    $\textbf{a}_t \sim \Pi_{\theta}(h_{\theta}(\textbf{s}_t))$ \\
                    $(\textbf{s}_{t + 1}, r_t) \sim \mathcal{P}(\textbf{s}_t, \textbf{a}_t), \mathcal{R}_{\theta}(\textbf{s}_t, \textbf{a}_t)$ \\
                    $\mathcal{D} \leftarrow \mathcal{D} \cup (\textbf{s}_t, \textbf{a}_t, r_t, \textbf{s}_{t + 1})$ \\
                }
                \For{step = 0, \dots, S}{
                    \textcolor{gray}{// $H$-step sampling from latent buffer} \\
                    $\{\textbf{s}_t, \textbf{a}_t, r_t, \textbf{s}_{t + 1}\}_{t : t+H}^{g} \sim \mathcal{D}$ {\small \textbf{for}} \textit{G} {\small groups} \\
                    $\mu^G, \sigma^G = \texttt{\small compute\_moments}(\textbf{a}_t)$ {\small (Eq. \ref{eq:traj_moments})} \\
                    $\textbf{z}_t = h_{\theta}(\textbf{s}_t)$ \textbf{if} $\textbf{s}_t$ {\small is the first observation} \\
                    \For{$i = t, \dots, t + H$}{
                        $\textbf{z}_{i + 1} = d_{\theta}(\textbf{z}_i, \textbf{a}_i)$ {\small (Eq. \ref{eq:latent_consistency})} \\
                        $\hat{r}_i = R_{\theta}(\textbf{z}_i, \textbf{a}_i)$ {\small (Eq. \ref{eq:reward_consistency})}\\
                        \textcolor{gray}{// group sampling \& policy constraint} \\
                        \For{$g = 1, \dots, G$ }{ 
                            $\widehat{\textbf{a}}_i^g \sim \pi_{\theta}(\textbf{z}_i)$ \\                        
                            $\varepsilon = (\widehat{\textbf{a}}_i^g - \mu^{G}) / \sigma^{G}$ \\
                            $\widehat{\textbf{a}}_i^g \leftarrow \texttt{\small{threshold}}(\widehat{\textbf{a}}_i^g, \varepsilon)$ {\small (Eq. \ref{eq:kl_loss})} \\
                            $\hat{q}_i^g = Q_{\theta}(\textbf{z}_i, \widehat{\textbf{a}}_i^g)$ {\small (Eq. \ref{eq:value_consistency})} \\
                        }
                        $A_i^g = \texttt{\small{softmax}}(\widehat{\textbf{q}}^g)$ {\small (Eq.~\ref{eq:weights})} \\
                        $\mathcal{L}_{\pi}^{(i)} = \frac{1}{G} \sum_{g=1}^{G} A_i^g \log \pi_\theta(\widehat{\textbf{a}}_i^g \mid \textbf{z}_i)$
                    }
                    $\mathcal{L}_\pi = \frac{1}{H} \sum_{i=t}^{t+H} \left[ \mathcal{L}_{\pi}^{(i)} + \beta \mathcal{L}_{\mathrm{KL}} \right]$ {\small (Eq. \ref{eq:policy_objective})} \\
                    $\theta \leftarrow \theta - \eta \nabla_{\theta} \mathcal{L}_{\pi}$
                }
            }
        }
    \end{normalsize}
\end{algorithm}

Following Theorem 1 in ~\cite{sikchi2022learning}, we assume that the nominal policy $\pi_k$ is obtained through approximate policy iteration, and the resulting planner policy at the $k^{th}$ iteration is denoted as $\mu_k = \pi_{H,k}$. Let the value approximation error be bounded as $\|Q_k - Q^{\pi_k}\|_\infty \leq \varepsilon_k$, the model approximation error be $\varepsilon_m = \max_{\textbf{s}, \textbf{a}} \mathbf{D}_{\mathrm{TV}}(\mathcal{P} \mid \mid \hat{\mathcal{P}})$ with $\mathbf{D}_{\mathrm{TV}}$ as total variation distance, and the planner sub-optimality be $\varepsilon_{p,k}$. Let the reward function be bounded as $r(\textbf{s}, \textbf{a}) \in [0, R_{\max}]$, and define the upper bound of the value function as $Q_{\max} = R_{\text{max}}/(1 - \gamma)$. Therefore, the planner's sub-optimal policy performance that satisfies the following uniform bound:
\begin{equation*}
    \limsup_{k \to \infty} \frac{1 - \gamma^H}{2} \left| V^* - V^{\mu_k} \right| \leq
\end{equation*}
\begin{equation*}
    \limsup_{k \to \infty} \left[ E(\varepsilon_m, H, \gamma) + \frac{\varepsilon_{p,k}}{2} + \frac{\gamma^H(1 + \gamma^2)}{(1 - \gamma)^2} \varepsilon_k \right],
\end{equation*}
where the model error constant $E$ is defined as:
\begin{equation*}
    E (\varepsilon_m, H, \gamma ) = R_{\max} \sum_{t=0}^{H-1} \gamma^t t \varepsilon_m + \gamma^H H \varepsilon_m V_{\max}.
\end{equation*}

Thus, with policies $\pi, \pi' \in \Pi$ and the reward's upper bound is $R_{\max}$, the policy divergence is lower bounded by the performance gap as the following expression:
\begin{equation*}
    \frac{(1 - \gamma)^2}{2 R_{\max}} \left| J^{\pi} - J^{\pi'} \right| \leq \max_{\textbf{s}} \mathbf{D}_{\text{TV}}\left( \pi' \mid \mid \pi \right)
    \vspace{3pt}
\end{equation*}

To ensure stable policy updates in high-dimensional and sensitive control environments, we impose a trust-region constraint on the policy learning process with Kullback–Leibler (KL) divergence between roll-out and reference policies. Specifically, we constrain the updated policy $\pi$ to remain close to a reference or prior policy $\pi_{\text{old}}$ by imposing the following divergence condition $\mathbf{D}_{\mathrm{KL}}(\pi \mid \mid \pi_{\text{old}}) \leq \varepsilon$, where $\varepsilon$ is a small adaptive threshold, enforcing a residual learning objective, encouraging the policy to improve upon the previous iteration while limiting drastic changes~\cite{peters2010relative, schulman2015trust}. Moreover, it mitigates the risk of destabilizing learned behaviors, critical in continuous control settings, especially for locomotion skills. 

In practice, we implement this constraint as a policy constraint loss $\mathcal{L}^{\text{PC}}$ under the Lagrangian version \cite{nair2020awac}:
\begin{equation}
    \mathcal{L}^{\text{PC}} = \max \left\{ \mathbf{D}_{\mathrm{KL}}(\pi \mid \mid \pi_{\text{old}}) - \varepsilon,\, 0 \right\},
    \label{eq:kl_loss}
\end{equation}

\subsubsection{Group Relative Policy Constraint}
We adopt and improve GRPO \cite{shao2024deepseekmath} to enhance action group-based explicit advantage refinement. Specifically, GRPO enhances entropy-regularized policy gradient methods by leveraging group-wise action comparisons, enabling the policy to learn from relative action preferences rather than relying on potentially noisy absolute value targets. In standard actor-critic methods, policy gradients are directly scaled by absolute $Q$-values or advantage estimates, which may be sensitive to reward scaling and value estimation errors. These limitations become especially pronounced in long-horizon tasks such as robotic locomotion. GRPO addresses this issue by constructing a relative preference distribution across sampled groups.

Formally, at each state $\textbf{s}$, a set of $G$ actions $\{\textbf{a}_1, \dots, \textbf{a}_G\}$ is sampled, and their $Q$-values $\{q_1, \dots, q_G\}$ are computed. These are used to define softmax-based advantage scores:
\begin{equation}
    A_{i}(\mathbf{q}) = \frac{\exp(q_{i} / \tau)}{\sum_{j=1}^G \exp(q_{j} / \tau)},
    \label{eq:weights}
\end{equation}
where $\mathbf{q} = Q_\theta(\textbf{s}, \textbf{a}_i)$ denotes the estimates, and $\tau$ is a temperature parameter and $0 \leq A_{i}(\cdot) \leq 1$ as its property.

\begin{figure*}[t]
    \vspace{5pt}
    \centering
    \includegraphics[width=1.0\linewidth]{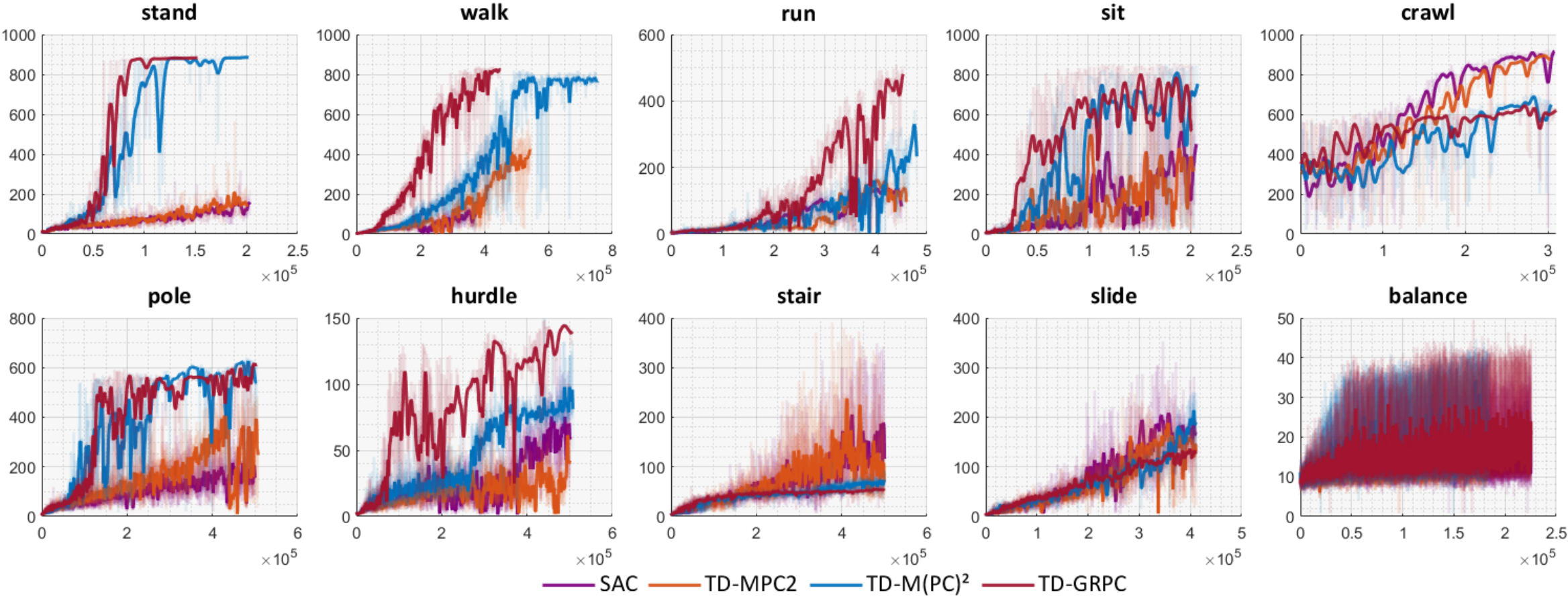}
    \vspace{-14pt}
    \caption{\textbf{Episode Returns of TD-GRPC and Baselines on H1–2 in Humanoid Locomotion Tasks:} TD-GRPC achieves rapid convergence over others in standing, walking, running, sitting, navigating through poles, hurdling, and sliding tasks, while it performs worse in crawling tasks. In general, TD-GRPC shows slightly better data-efficiency than TD-M(PC)$^2$ and significantly better sampling-efficiency than SAC and TD-MPC2, alongside the fact that TD-GRPC outperforms TD-MPC2 and SAC on many tasks quantitatively. Nevertheless, all benchmarked algorithms fail to accomplish more challenging tasks, such as stair-climbing and balancing on a ball-board platform.}
    \vspace{-14pt}
    \label{fig:returns_comparison}
\end{figure*}

As $\{\textbf{a}_i\}_{i=1}^G$ is a group of $G$ actions sampled from a policy $\pi_\theta(\textbf{s})$ at the state $\textbf{s}$ with $r_i = r_{\theta}(\textbf{s}, \textbf{a}_i)$ and $q_i = Q_{\theta}(\textbf{s}, \textbf{a}_i)$ are the reward and estimated value for each action, respectively, we assume that  $||\nabla_\theta \log \pi_\theta(\textbf{a} \mid \textbf{s})|| = C$, and $q_i$, $r_i$ are bounded above with $\forall \textbf{a} \in \mathcal{A}$ and $\forall \textbf{s} \in \mathcal{S}$. We obtain:
\begin{equation}
    \mathrm{Var} \left[ \nabla_{\theta} \mathcal{L}_{\text{softmax}} \right] \leq \mathrm{Var} \left[ \nabla_{\theta} \mathcal{L}_{\text{std-norm}} \right],
    \label{eq:variance_inequality}
\end{equation}
with $\mathcal{L}_{\text{softmax}}$ and $\mathcal{L}_{\text{std-norm}}$ are the softmax-based and standard normalized advantage scores, respectively. Additionally, the normalized scores are unbounded, but the advantage scores are bounded in the range of 0 to 1, the variance of gradient of $\mathcal{L}_{\text{softmax}}$ is thus smaller than that of $\mathcal{L}_{\text{std-norm}}$:
\begin{equation}
    ||\nabla_{\theta} \mathcal{L}_{\text{softmax}}|| \text{ is bounded, }  ||\nabla_{\theta} \mathcal{L}_{\text{std-norm}}||\text{ is unbounded} 
    \label{eq:bounded_loss}
\end{equation}
yields more stable policy updates at some constant $C$ that asymptotically bounds $||\nabla_{\theta} \log \pi_\theta(\textbf{a} \mid \textbf{s})||$. Two keys favor softmax-based over normalized advantages. First, their outputs lie between $0$ and $1$, limiting the impact of outliers. Meanwhile, normalized advantages induce large magnitudes under noise, leading to high-variance gradients. Second, policy gradients scale with the advantage values. If the advantage is very large or small, the gradient steps are disproportionately unstable. Therefore, softmax-based advantages smooth out extreme values and act like a soft attention mechanism, giving more stable updates.

With the group relative weights in Eq. \ref{eq:weights} and based on Eq. \ref{eq:variance_inequality} and Eq. \ref{eq:bounded_loss}, the improved GRPO objective is defined as:
\begin{equation}
    \mathcal{L}_{\pi}^{\text{GRPO}}(\theta) = \frac{1}{G} \sum_{i=1}^{G} A_{i}(\textbf{q}) \log \pi_{\theta}(\textbf{a}_i \mid \textbf{s})
    \label{eq:grpo}
\end{equation}
where $\mu_k$ denotes the behavior policy at $k^{th}$ iteration from the buffer $\mathcal{D}$ obtained from Eq. \ref{eq:traj_moments}. The KL constraint ensures the updated policy remains within a trust region of $\pi$. The overall policy objective combines Eq. \ref{eq:kl_loss} with Eq. \ref{eq:grpo}:
\begin{equation}
    \mathcal{L}_{\pi}(\theta) = \underbrace{ \frac{1}{G} \sum_{i=1}^{G} A_{i}(\textbf{q}) \log \pi_{\theta}(\textbf{a}_i \mid \textbf{s})) }_{\textcolor{gray}{\text{improved GRPO}}} + \beta \underbrace{ \log \mu(\textbf{a} \mid \textbf{s}) }_{\textcolor{gray}{\text{policy constraint}}},
    \label{eq:policy_objective}
\end{equation}
where $\beta$ is a weighting coefficient controlling the penalty strength. The second term of Eq. \ref{eq:policy_objective} imposes a residual-style regularization equivalent to the trust-region \cite{schulman2015trust}.

Meanwhile, the latent dynamics $d_{\theta}$, encoder $h_{\theta}$, reward network $R_{\theta}$, and value network $Q_{\theta}$ are concurrently optimized by the following model objective:
\begin{subequations}
    \begin{align}
        \mathcal{L}&(\theta;\text{ }\Gamma_{i}) = 
        \left\| d_\theta(\mathbf{z}_i, \mathbf{a}_i) - h_\theta(\mathbf{s}_{i+1}) \right\|_2^2 \label{eq:latent_consistency} \\
        &+ \left\| R_\theta(\mathbf{z}_i, \mathbf{a}_i) - r_i \right\|_2^2 \label{eq:reward_consistency} \\
        &+ \left\| Q_\theta(\mathbf{z}_i, \mathbf{a}_i) - \left[r_i + \gamma Q_\theta(\mathbf{z}_{i+1}, \pi_\theta(\mathbf{z}_{i+1}))\right] \right\|_2^2 \label{eq:value_consistency}
    \end{align}
    \label{eq:model_objective}
    \vspace{-12pt}
\end{subequations}

The training procedure with temporal difference learning and group relative policy constraints is summarized in Alg. \ref{alg:td_grpc} along with the pipeline shown in Fig. \ref{fig:tdgrpc_framework}. Meanwhile, the inference process remains the same as in TD-MPC \cite{hansen2022temporal} with cross-entropy method \cite{rubinstein1997optimization} on rolled-out learned policies.

%% file: 04_evaluation.tex
\section{Experimental Results \& Analysis}
\label{sec:evaluations}

We evaluate our proposed method on \texttt{HumanoidBench} \cite{sferrazza2024humanoidbench} with $10$ locomotion tasks, including standing, walking, running, sitting on a chair, crawling under a tunnel, navigating through poles while avoiding collision, hurdling, climbing stairs, sliding, and balancing on a ball-board platform, on \textbf{the 26-DoF Unitree H1-2 humanoid} with two legs ($6$-DoF each $\times 2$) and two arms ($7$-DoF each $\times 2$), which is more dynamically flexible compared to H1 with 21 DoFs. While training, we also include the hands in the robot's model to ensure that the learned policies consider both hands' mass, although the robot does not encounter manipulation tasks. Additionally, H1-2 weighs about $70$ kg compared to H1’s $47$ kg, presenting challenges for learning body dynamics due to its significantly heavier build. For comparison, we select the following three state-of-the-art RL methods as baselines:
\begin{itemize}
    \item \textbf{Soft Actor-Critic (SAC)} \cite{haarnoja2018soft}: a model-free, off-policy RL algorithm with maximum entropy RL \cite{ziebart2008maximum}.
    \item \textbf{TD-MPC2} \cite{hansen2023td}: the state-of-the-art MBRL that combines MPC with TD learning for diverse continuous control tasks in the DeepMind control suite \cite{tassa2018deepmind}.
    \item \textbf{TD-M(PC)$^2$} \cite{lin2025td}: the most recent variant of TD-MPC2 for humanoid locomotion that utilizes KL-regularized policy learning to overcome value overestimation.
    \vspace{-15pt}
\end{itemize}.

\begin{figure*}[t]
    \vspace{4pt}
    \centering
    \includegraphics[width=1.0\linewidth]{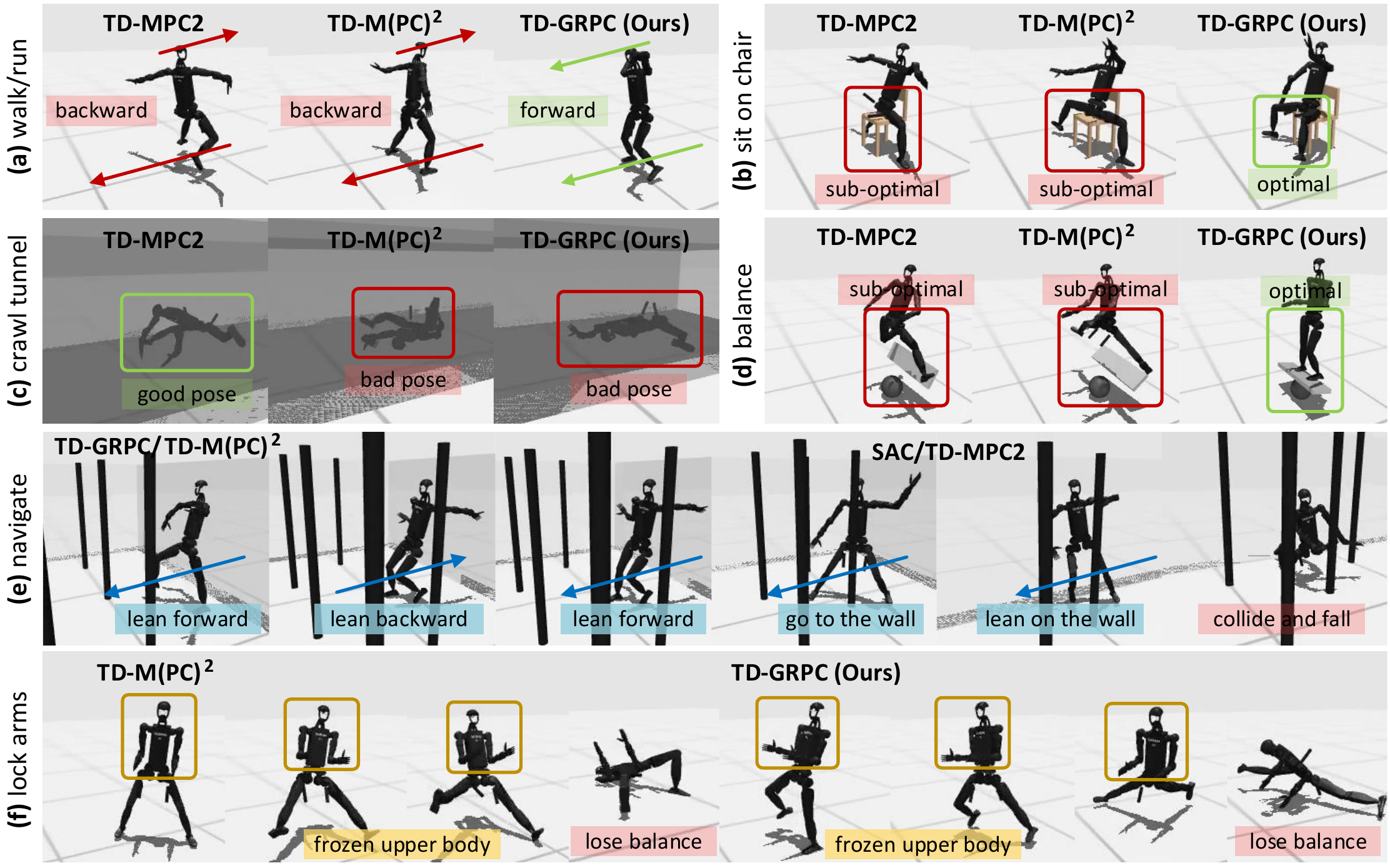}
    \caption{\textbf{Behavioral Analysis of H1-2 in Humanoid Locomotion Tasks:} (a) \textit{Walking and Running Backwards:} walking direction comparisons between TD-MPC2, TD-M(PC)$^2$, and our approach. TD-GRPC directs H1-2 to move forward, but TD-MPC2 and TD-M(PC)$^2$ make it walk/run backward. (b) \textit{Sitting Stability:} sitting pose comparisons between TD-MPC2, TD-M(PC)$^2$, and TD-GRPC. TD-GRPC achieves optimal sitting leg pose. All frames are taken from the last state of the evaluation episode. (c) \textit{Crawling Pose and Height:} The crawling pose and height produced by TD-MPC2 are better than those generated by TD-M(PC)$^2$ and TD-GRPC. (d) \textit{Balancing is Hard for Heavy Body:} Due to its heavy body weight, all three methods suffer difficulties in keeping H1-2 balance itself on the platform. However, TD-GRPC can balance the robot for a short period; meanwhile, TD-MPC2 and TD-M(PC)$^2$ make the robot's legs flick the board away and fail immediately. (e) \textit{Navigating through Standing Poles:} TD-M(PC)$^2$ and TD-GRPC induce a standing behavior without navigating, while SAC and TD-MPC2 produce valid motions but collide with poles and fail. (f) \textit{Arm-Balancing Helps Locomotion:} Freezing the upper body of H1-2 makes walking and running unstable. This experimental finding is evaluated with both TD-M(PC)$^2$ and TD-GRPC.}
    \vspace{-16pt}
    \label{fig:behavioral_analysis}
\end{figure*}

\subsection{Quantitative Comparisons}

As shown in Fig. \ref{fig:returns_comparison}, we compare the episode returns TD-GRPC against the baselines on the humanoid locomotion tasks mentioned. All algorithms are run with the planning horizon $H$ of $3$, size of latent buffer $\mathcal{D}$ of $1,000,000$, discount factor $\gamma$ of $0.995$, and learning rate of $0.0003$ on an AMD Ryzen 9 7950X3D CPU and an NVIDIA RTX 4090 GPU. For TD-GRPC, we set the number of groups $G$ as $3$.

\subsubsection{Standing} 
TD-GRPC effectively trains H1-2 to stand upright and stably faster than TD-MPC2 and TD-M(PC)$^2$. Both TD-GRPC and TD-M(PC)$^2$ surpass the reward of $800$. Meanwhile, SAC and TD-MPC2 fail to teach the robot to stand within $200,000$ iterations.

\subsubsection{Walking} 
As with the standing task, H1-2, when trained with TD-GRPC, can walk more rapidly than other baselines. Both TD-M(PC)$^2$ and TD-GRPC gain the reward of more than $750$ within $500,000$ iterations, but TD-GRPC allows the robot to walk at $400,000^{th}$ iteration. SAC and TD-MPC fail to enable the robot to walk during the same training period. However, the walking behavior among the algorithms is further analyzed in Sec. \ref{sec:qualitative_results_walking_running}.

\begin{table*}[t]
    \vspace{4pt}
    \centering
    \caption{Solving ability of SAC \cite{haarnoja2018soft}, TD-MPC2 \cite{hansen2023td}, TD-M(PC)$^2$ \cite{lin2025td}, and our proposed method, TD-GRPC, of locomotion tasks on H1-2 in \texttt{HumanoidBench} \cite{sferrazza2024humanoidbench}:~\cmark~for tasks that are solved sufficiently,~\omark~for tasks that need additional mild refinements for success, and~\xmark~for tasks that need further intensive policy-constraint learning of whole-body and selective environmental dynamics features.}
    \vspace{-2pt}
    \begin{tabular}{l ccccccccccc}
        \toprule
        \multirow{2}{*}{\diagbox{Method}{Task}} & \multicolumn{10}{c}{\makecell{Locomotion Tasks in \texttt{HumanoidBench} Environment \cite{sferrazza2024humanoidbench}}} \\
        \cmidrule(lr){2-11}
        & \texttt{stand} & \texttt{walk} & \texttt{run} & \texttt{sit} & \texttt{crawl} & \texttt{pole} & \texttt{hurdle} & \texttt{stair} & \texttt{slide} & \texttt{balance} \\
        \midrule \midrule
        SAC \cite{haarnoja2018soft} & \xmark & \xmark & \xmark & \xmark & \cmark & \xmark & \xmark & \xmark & \omark & \xmark \\
        TD-MPC2 \cite{hansen2023td} & \xmark & \xmark & \xmark & \omark & \cmark & \xmark & \xmark & \xmark & \omark & \xmark \\
        TD-M(PC)$^2$ \cite{lin2025td} & \cmark & \omark & \omark & \omark & \omark & \omark & \xmark & \xmark & \omark & \xmark \\
        \midrule
        \textbf{TD-GRPC (Ours)} & \cmark & \cmark & \cmark & \cmark & \omark & \omark & \omark & \xmark & \omark & \omark \\
        \bottomrule
    \end{tabular}
    \label{tab:solving_ability}
    \vspace{-16pt}
\end{table*}

\subsubsection{Running} 
The running task is defined analogously to the walking task, but with a goal speed higher than walking speed. While SAC and TD-MPC2 fail to generate proper running policies in $450,000$ iterations, TD-M(PC)$^2$ and TD-GRPC successfully enable the H1-2 to run, but TD-GRPC converges faster than TD-M(PC)$^2$ does. The illustrative behavior for this task is presented in Sec. \ref{sec:qualitative_results_walking_running}.

\subsubsection{Sitting} 
Similar to the standing and walking tasks, TD-GRPC continues showing faster convergence than SAC, TD-MPC2, and TD-M(PC)$^2$ in learning to sit on the chair. Unlike other algorithms, the experiments also confirm that the robot can sit steadily in a good pose with TD-GRPC, as described in Sec. \ref{sec:qualitative_results_sitting}. Both TD-M(PC)$^2$ and TD-GRPC-generated policies exceed the reward of $600$ within $200,000$ iterations. 

\subsubsection{Crawling} 
SAC and TD-MPC2, on the other hand, show their superior capabilities to get the robot to crawl through the tunnel with the rewards of over $800$ within $300,000$ iterations. TD-M(PC)$^2$ and TD-GRPC get the robot to crawl, but not enough to pass to the other end of the tunnel; their poses are visually explained in Sec. \ref{sec:qualitative_results_crawling}. 

\subsubsection{Balancing} The reward learned by TD-MPC2, TD-M(PC)$^2$, and TD-GRPC are approximately similar on the H1-2. The same phenomenon applies to SAC and TD-MPC2. This experiment is considered a hard task for H1-2 due to its heavy weight; the behavior is provided in Sec. \ref{sec:qualitative_results_balancing}.

\subsubsection{Navigating through Standing Poles} 
The rewards gained by TD-M(PC)$^2$ and TD-GRPC are higher than those achieved by SAC and TD-MPC. Within $500,000$ training iterations, TD-GRPC again shows its convergence well beyond TD-M(PC)$^2$, alluding that the task is concluded. Nevertheless, the robot's behaviors from TD-M(PC)$^2$ and TD-GRPC are distinct from those performed by SAC and TD-MPC2, as studied in Sec. \ref{sec:qualitative_results_navigating}.

\subsubsection{Stair-Climbing} 
Learning whole-body dynamics is intricate for all baseline methods and our proposed method in stair-climbing tasks. Unlike walking or running, climbing stairs requires more than locomotion on an even surface; the robot has to change its foot height iteratively during stepping up. All methods struggle at a reward value of $200$.

\subsubsection{Hurdling} 
Hurdling is a running variant; however, the robot must jump over the tracks while running. During this task, TD-GRPC can let the robot jump over one track without collision, while others generate sub-optimal actions and fail. Evidently, TD-GRPC promotes the robot to learn as it produces higher learning reward curves than other methods, surpassing the reward threshold of $100$.

\subsubsection{Sliding} 
Similar to the stair-climbing objective, sliding requires the robot to climb up a hill-like landscape without needing foot-stepping behavior. All benchmarked methods and TD-GRPC are able to learn this task at the reward of $200$, and generate physically-meaning actions (\textit{i.e.}, knee-walking) to achieve the task's goal. However, the robot can only go up one hill and not go to other hills.

\subsection{Qualitative Comparisons \& Behavioral Analysis}
\label{sec:qualitative_results}
Not just upon task completion, we outline selected humanoid locomotion tasks and analyze our findings on the robot's behavior as lessons learned for humanoid locomotion.

\subsubsection{Walking \& Running Backwards}
\label{sec:qualitative_results_walking_running}
While TD-MPC2 and TD-M(PC)$^2$ perform well on H1, they fail to correct walking and running poses on the H1-2, where they both make the robot walk or run backward with its head looking in the opposite direction. SAC is neither able to run nor walk the H1-2 properly. TD-GRPC successfully enables forward locomotion for walking and running, as shown in Fig. \ref{fig:behavioral_analysis}a.

\subsubsection{Sitting Stability} 
\label{sec:qualitative_results_sitting}
The robot is able to sit stably on the chair with limited jerky motions while training with TD-GRPC, but not TD-MPC2 and TD-M(PC)$^2$. In addition, the leg poses are learned to be put appropriately when sitting, as shown in Fig. \ref{fig:behavioral_analysis}b. While trained with SAC and TD-MPC2, the robot can not sit stably on the chair.

\subsubsection{Crawling Pose \& Height}
\label{sec:qualitative_results_crawling}
Righteously crawling with proper poses is more challenging for TD-M(PC)$^2$ and TD-GRPC. As shown in Fig. \ref{fig:behavioral_analysis}c, they both stuck at sub-optimal poses and could not crawl to the other end of the tunnel, failing to complete the tasks. On the contrary, TD-MPC2 completes its tasks with good body posture and head height.

\subsubsection{Balancing is Hard for Heavy Body}
\label{sec:qualitative_results_balancing}
Compared to H1, which is lighter, H1-2 suffers from different body dynamics to keep itself balanced on the ball-board platform. We find that TD-GRPC allows the robot to balance for a period of time before failing. In contrast, TD-M(PC)$^2$ and TD-MPC2 fail to generate physical behavior for balancing from the beginning: their legs flick the board away instead of standing on it, as illustrated in Fig. \ref{fig:behavioral_analysis}d.

\subsubsection{Navigation through Standing Poles}
\label{sec:qualitative_results_navigating}
Despite the high-return rewards of approximately 600 from TD-M(PC)$^2$ and TD-GRPC, the robot's behavior through these algorithms differs from when learning with SAC and TD-MPC2, where the task requires the robot to move forward while avoiding collision with the poles to gain rewards. Meanwhile, the policies trained on SAC and TD-MPC2 tell the robot to go to the side of the room and follow the room edge, effectively avoiding collision as a safe strategy. However, the robot collides with the poles and falls, unable to accomplish the task. TD-M(PC)$^2$ and TD-GRPC direct the robot to move forward, go back, and repeat such actions until the episode ends. As a result, the robot does not move much, staying at the same spot, but is still gaining rewards (Fig. \ref{fig:behavioral_analysis}e).

\subsubsection{Arm-Balancing Helps Locomotion} 
Throughout locomotion experiments on \texttt{HumanoidBench}, we observe that the robot arms are ``\textit{ill-posed}'' and not in optimal states. While investigating such behaviors, we find that arm-balancing helps humanoid locomotion despite random arm movements while executing tasks. For instance, we lock both arms at fixed positions, and make it learn a locomotion task (\textit{e.g.}, walking, running). Consequently, the robot could not accomplish the tasks when learning with different algorithms. Fig. \ref{fig:behavioral_analysis}f reflects their intricacy and barely complete the tasks, where the robots unsteadily fall during their learning episodes. Constraining these arms' poses could further benefit robot actions in real-world scenarios.

\subsection{Demonstration}
Besides solving abilities of algorithms learned from quantitative and qualitative results in Tab. \ref{tab:solving_ability}, the demonstration video of our experiments can be seen in the supplementary document for comparisons of H1-2's performance across locomotion tasks between benchmarked algorithms.

%% file: 05_conclusions.tex
\section{Conclusions}

We presented TD-GRPC -- a framework incorporating GRPO, explicit policy constraints, and TD learning for stable policy updates during learning humanoid locomotion tasks. Via this RL-based method, we achieve robust and sample-efficient learning by constraining policy rollouts in latent space without restricting planner flexibility. Our results on \texttt{HumanoidBench} with the 26-DoF Unitree H1-2 humanoid demonstrate that TD-GRPC surpasses existing baselines in stability and quantitative and qualitative performances and sets a foundation for scalable, constraint-aware RL in high-dimensional complex humanoid control.

%% file: 00_main.bbl
\begin{thebibliography}{10}
\providecommand{\url}[1]{#1}
\csname url@samestyle\endcsname
\providecommand{\newblock}{\relax}
\providecommand{\bibinfo}[2]{#2}
\providecommand{\BIBentrySTDinterwordspacing}{\spaceskip=0pt\relax}
\providecommand{\BIBentryALTinterwordstretchfactor}{4}
\providecommand{\BIBentryALTinterwordspacing}{\spaceskip=\fontdimen2\font plus
\BIBentryALTinterwordstretchfactor\fontdimen3\font minus \fontdimen4\font\relax}
\providecommand{\BIBforeignlanguage}[2]{{%
\expandafter\ifx\csname l@#1\endcsname\relax
\typeout{** WARNING: IEEEtran.bst: No hyphenation pattern has been}%
\typeout{** loaded for the language `#1'. Using the pattern for}%
\typeout{** the default language instead.}%
\else
\language=\csname l@#1\endcsname
\fi
#2}}
\providecommand{\BIBdecl}{\relax}
\BIBdecl

\bibitem{radosavovic2023learning}
I.~Radosavovic, T.~Xiao, B.~Zhang, T.~Darrell, J.~Malik, and K.~Sreenath, ``Learning humanoid locomotion with transformers,'' \emph{CoRR}, 2023.

\bibitem{radosavovic2024real}
------, ``Real-world humanoid locomotion with reinforcement learning,'' \emph{Science Robotics}, vol.~9, no.~89, p. eadi9579, 2024.

\bibitem{nagabandi2018neural}
A.~Nagabandi, G.~Kahn, R.~S. Fearing, and S.~Levine, ``Neural network dynamics for model-based deep reinforcement learning with model-free fine-tuning,'' in \emph{IEEE ICRA}, 2018.

\bibitem{chua2018deep}
K.~Chua, R.~Calandra, R.~McAllister, and S.~Levine, ``Deep reinforcement learning in a handful of trials using probabilistic dynamics models,'' \emph{NeurIPS}, 2018.

\bibitem{argenson2020model}
A.~Argenson and G.~Dulac-Arnold, ``Model-based offline planning,'' \emph{arXiv preprint arXiv:2008.05556}, 2020.

\bibitem{kouvaritakis2016model}
B.~Kouvaritakis and M.~Cannon, ``Model predictive control,'' \emph{Switzerland: Springer International Publishing}, vol.~38, no. 13-56, p.~7, 2016.

\bibitem{haarnoja2018soft}
T.~Haarnoja, A.~Zhou, P.~Abbeel, and S.~Levine, ``Soft actor-critic: Off-policy maximum entropy deep reinforcement learning with a stochastic actor,'' in \emph{ICML}, 2018.

\bibitem{schrittwieser2020mastering}
J.~Schrittwieser, I.~Antonoglou, T.~Hubert, K.~Simonyan, L.~Sifre, S.~Schmitt, A.~Guez, E.~Lockhart, D.~Hassabis, T.~Graepel \emph{et~al.}, ``Mastering atari, go, chess and shogi by planning with a learned model,'' \emph{Nature}, vol. 588, no. 7839, pp. 604--609, 2020.

\bibitem{ye2021mastering}
W.~Ye, S.~Liu, T.~Kurutach, P.~Abbeel, and Y.~Gao, ``Mastering atari games with limited data,'' \emph{NeurIPS}, 2021.

\bibitem{sikchi2022learning}
H.~Sikchi, W.~Zhou, and D.~Held, ``Learning off-policy with online planning,'' in \emph{CoRL}, 2022.

\bibitem{hansen2023td}
N.~Hansen, H.~Su, and X.~Wang, ``Td-mpc2: Scalable, robust world models for continuous control,'' \emph{arXiv preprint arXiv:2310.16828}, 2023.

\bibitem{lin2025td}
H.~Lin, P.~Wang, J.~Schneider, and G.~Shi, ``Improving temporal difference mpc through policy constraint,'' \emph{arXiv preprint arXiv:2502.03550}, 2025.

\bibitem{sutton1991dyna}
R.~S. Sutton, ``Dyna, an integrated architecture for learning, planning, and reacting,'' \emph{ACM Sigart Bulletin}, vol.~2, no.~4, pp. 160--163, 1991.

\bibitem{janner2019trust}
M.~Janner, J.~Fu, M.~Zhang, and S.~Levine, ``When to trust your model: Model-based policy optimization,'' \emph{NeurIPS}, 2019.

\bibitem{kaiser2019model}
L.~Kaiser, M.~Babaeizadeh, P.~Milos, B.~Osinski, R.~H. Campbell, K.~Czechowski, D.~Erhan, C.~Finn, P.~Kozakowski, S.~Levine \emph{et~al.}, ``Model-based reinforcement learning for atari,'' \emph{arXiv preprint arXiv:1903.00374}, 2019.

\bibitem{pham2025pay}
T.~Pham and A.~Cangelosi, ``Pay attention to what and where? interpretable feature extractor in vision-based deep reinforcement learning,'' in \emph{IJCNN}, 2025.

\bibitem{hwangbo2019learning}
J.~Hwangbo, J.~Lee, A.~Dosovitskiy, D.~Bellicoso, V.~Tsounis, V.~Koltun, and M.~Hutter, ``Learning agile and dynamic motor skills for legged robots,'' \emph{Science Robotics}, vol.~4, no.~26, p. eaau5872, 2019.

\bibitem{lee2020learning}
J.~Lee, J.~Hwangbo, L.~Wellhausen, V.~Koltun, and M.~Hutter, ``Learning quadrupedal locomotion over challenging terrain,'' \emph{Science robotics}, vol.~5, no.~47, p. eabc5986, 2020.

\bibitem{miki2022learning}
T.~Miki, J.~Lee, J.~Hwangbo, L.~Wellhausen, V.~Koltun, and M.~Hutter, ``Learning robust perceptive locomotion for quadrupedal robots in the wild,'' \emph{Science robotics}, vol.~7, no.~62, p. eabk2822, 2022.

\bibitem{hafner2019dream}
D.~Hafner, T.~Lillicrap, J.~Ba, and M.~Norouzi, ``Dream to control: Learning behaviors by latent imagination,'' \emph{arXiv preprint arXiv:1912.01603}, 2019.

\bibitem{hafner2020mastering}
D.~Hafner, T.~Lillicrap, M.~Norouzi, and J.~Ba, ``Mastering atari with discrete world models,'' \emph{arXiv preprint arXiv:2010.02193}, 2020.

\bibitem{hafner2023mastering}
D.~Hafner, J.~Pasukonis, J.~Ba, and T.~Lillicrap, ``Mastering diverse domains through world models,'' \emph{arXiv preprint arXiv:2301.04104}, 2023.

\bibitem{ha2018recurrent}
D.~Ha and J.~Schmidhuber, ``Recurrent world models facilitate policy evolution,'' \emph{NeurIPS}, 2018.

\bibitem{lambert2020objective}
N.~Lambert, B.~Amos, O.~Yadan, and R.~Calandra, ``Objective mismatch in model-based reinforcement learning,'' \emph{arXiv preprint arXiv:2002.04523}, 2020.

\bibitem{xu2022feasibility}
Y.~Xu, N.~Hansen, Z.~Wang, Y.-C. Chan, H.~Su, and Z.~Tu, ``On the feasibility of cross-task transfer with model-based reinforcement learning,'' \emph{arXiv preprint arXiv:2210.10763}, 2022.

\bibitem{hansen2022temporal}
N.~Hansen, X.~Wang, and H.~Su, ``Temporal difference learning for model predictive control,'' \emph{arXiv preprint arXiv:2203.04955}, 2022.

\bibitem{clavera2020model}
I.~Clavera, V.~Fu, and P.~Abbeel, ``Model-augmented actor-critic: Backpropagating through paths,'' \emph{arXiv preprint arXiv:2005.08068}, 2020.

\bibitem{fujimoto2019off}
S.~Fujimoto, D.~Meger, and D.~Precup, ``Off-policy deep reinforcement learning without exploration,'' in \emph{ICML}, 2019.

\bibitem{kostrikov2021offline}
I.~Kostrikov, A.~Nair, and S.~Levine, ``Offline reinforcement learning with implicit q-learning,'' \emph{arXiv preprint arXiv:2110.06169}, 2021.

\bibitem{bellemare2017distributional}
M.~G. Bellemare, W.~Dabney, and R.~Munos, ``A distributional perspective on reinforcement learning,'' in \emph{ICML}, 2017.

\bibitem{levine2020offline}
S.~Levine, A.~Kumar, G.~Tucker, and J.~Fu, ``Offline reinforcement learning: Tutorial, review, and perspectives on open problems,'' \emph{arXiv preprint arXiv:2005.01643}, 2020.

\bibitem{shao2024deepseekmath}
Z.~Shao, P.~Wang, Q.~Zhu, R.~Xu, J.~Song, X.~Bi, H.~Zhang, M.~Zhang, Y.~Li, Y.~Wu \emph{et~al.}, ``Deepseekmath: Pushing the limits of mathematical reasoning in open language models,'' \emph{arXiv preprint arXiv:2402.03300}, 2024.

\bibitem{sferrazza2024humanoidbench}
C.~Sferrazza, D.-M. Huang, X.~Lin, Y.~Lee, and P.~Abbeel, ``Humanoidbench: Simulated humanoid benchmark for whole-body locomotion and manipulation,'' \emph{arXiv preprint arXiv:2403.10506}, 2024.

\bibitem{peters2003reinforcement}
J.~Peters, S.~Vijayakumar, and S.~Schaal, ``Reinforcement learning for humanoid robotics,'' in \emph{IEEE-RAS Humanoids}, 2003.

\bibitem{kumar2019stabilizing}
A.~Kumar, J.~Fu, M.~Soh, G.~Tucker, and S.~Levine, ``Stabilizing off-policy q-learning via bootstrapping error reduction,'' \emph{NeurIPS}, vol.~32, 2019.

\bibitem{kumar2020conservative}
A.~Kumar, A.~Zhou, G.~Tucker, and S.~Levine, ``Conservative q-learning for offline reinforcement learning,'' \emph{NeurIPS}, vol.~33, pp. 1179--1191, 2020.

\bibitem{fujimoto2021minimalist}
S.~Fujimoto and S.~S. Gu, ``A minimalist approach to offline reinforcement learning,'' \emph{NeurIPS}, vol.~34, pp. 20\,132--20\,145, 2021.

\bibitem{peng2019advantage}
X.~B. Peng, A.~Kumar, G.~Zhang, and S.~Levine, ``Advantage-weighted regression: Simple and scalable off-policy reinforcement learning,'' \emph{arXiv preprint arXiv:1910.00177}, 2019.

\bibitem{garg2023extreme}
D.~Garg, J.~Hejna, M.~Geist, and S.~Ermon, ``Extreme q-learning: Maxent rl without entropy,'' \emph{arXiv preprint arXiv:2301.02328}, 2023.

\bibitem{williams2016aggressive}
G.~Williams, P.~Drews, B.~Goldfain, J.~M. Rehg, and E.~A. Theodorou, ``Aggressive driving with model predictive path integral control,'' in \emph{IEEE ICRA}, 2016.

\bibitem{singh1994upper}
S.~P. Singh and R.~C. Yee, ``An upper bound on the loss from approximate optimal-value functions,'' \emph{Machine Learning}, vol.~16, pp. 227--233, 1994.

\bibitem{peters2010relative}
J.~Peters, K.~Mulling, and Y.~Altun, ``Relative entropy policy search,'' in \emph{Proceedings of the AAAI Conference on Artificial Intelligence}, vol.~24, no.~1, 2010, pp. 1607--1612.

\bibitem{schulman2015trust}
J.~Schulman, S.~Levine, P.~Abbeel, M.~Jordan, and P.~Moritz, ``Trust region policy optimization,'' in \emph{ICML}, 2015.

\bibitem{nair2020awac}
A.~Nair, A.~Gupta, M.~Dalal, and S.~Levine, ``Awac: Accelerating online reinforcement learning with offline datasets,'' \emph{arXiv preprint arXiv:2006.09359}, 2020.

\bibitem{rubinstein1997optimization}
R.~Y. Rubinstein, ``Optimization of computer simulation models with rare events,'' \emph{European Journal of Operational Research}, vol.~99, no.~1, pp. 89--112, 1997.

\bibitem{ziebart2008maximum}
B.~D. Ziebart, A.~L. Maas, J.~A. Bagnell, A.~K. Dey \emph{et~al.}, ``Maximum entropy inverse reinforcement learning.'' in \emph{Aaai}, vol.~8.\hskip 1em plus 0.5em minus 0.4em\relax Chicago, IL, USA, 2008, pp. 1433--1438.

\bibitem{tassa2018deepmind}
Y.~Tassa, Y.~Doron, A.~Muldal, T.~Erez, Y.~Li, D.~d.~L. Casas, D.~Budden, A.~Abdolmaleki, J.~Merel, A.~Lefrancq \emph{et~al.}, ``Deepmind control suite,'' \emph{arXiv preprint arXiv:1801.00690}, 2018.

\end{thebibliography}
